# Hybrid Ensemble for Fake News Detection: An attempt

Lovedeep Singh[1]

[1]Punjab Engineering College, Chandigarh, India
`masterlovedeep.singh97@gmail.com`

**Abstract.** Fake News Detection has been a challenging problem in the field of Machine Learning. Researchers have approached it via several techniques using old Statistical Classification models and modern Deep Learning. Today, with the growing amount of data, developments in the field of NLP and ML, and an increase in the computation power at disposal, there are infinite permutations and combinations to approach this problem from a different perspective. In this paper, we try different methods to tackle Fake News, and try to build, and propose the possibilities of a Hybrid Ensemble combining the classical Machine Learning techniques with the modern Deep Learning Approaches

## 1 Introduction

Fake News spreads through a variety of mediums in many forms. It can spread via word of mouth in the form of audio or the form of images via social media platforms like Facebook, WhatsApp, and Twitter. One of the primitive forms of fake news has been text. This paper discusses possibilities to tackle this category and is an attempt towards a hybrid ensemble for fake news detection with the form of fake news being textual.

## 2 Related Work

Ensemble techniques[1] appear promising in the field of machine learning. The ideal goal of an ensemble technique is to incorporate the positive qualities of different models while avoiding the bias present in a particular model. People have tried various approaches to classify text using machine learning methods on top of a preprocessed data, preprocessing done via various NLP techniques[2]. Researchers have utilized classification models such as SVM, KNN, Logistic Regression etc. as well as experimented with deep learning techniques[3]. They have also tried to combine algorithms sequentially or parallelly[4] to improve the accuracy of the models.

## 3 Proposed Method

### 3.1 Dataset

There are some good datasets available for the detection of fake news. We use the LIAR[5] dataset in the training and testing of our model. The dataset classifies the news into six categories, three bending toward truth, three bending towards false. We make this categorization broader by combining three positive categories as true and the three negative categories as false with the hope that with more data in each category, the models will identify more differences and perform better. The dataset had in actual 12 columns for each label. We use only selected columns and introduce some new entries to the columns.

### 3.2 Preprocessing

While it makes more sense to use all the columns present in the dataset, we do not use them. Using all columns in building a machine learning model may improve the performance of the model but will also introduce a bias in the model for a specific type of dataset. While we cannot totally sway away from the bias introduced by the dataset's political nature, we can avoid creating dependency on a large number of unique features that are not available in many practical scenarios. In practical life, whenever we encounter a piece of news, and we are not sure whether the news is genuine or fake, it is because we do not have enough parameters to make a sound conclusion We want our machine to find those elementary features in case they exist, which can help us say with some certainty whether the news is fake or genuine.

### 3.3 Removal of Columns

The dataset initially contained ['ID', 'label', 'statement', 'subject(s)', 'speaker', 'speaker's job title', 'state info', 'party affiliation', 'barely true counts', 'false counts', 'half true counts', 'mostly true counts', 'pants on fire', 'venue/location']. We keep only two columns label and statement. After removing other columns, we are left with ['label', 'statement'].

### 3.4 Addition of new columns

We add some new features that are not specific to any domain or structure of the dataset. The only requirement is that the data present is in textual form. The intent is to try and figure out the intricate differences between fake and true news independent of the domain. We want to answer questions that are common in all domains. These are the features that are captured in the Linguistic aspect of the language being used to convey the information rather than in the domain knowledge available to an expert. There are a whole lot of linguistic features available, or one can think of. We limit ourselves to the following features, namely Readability[1] (ease of understanding of the text), CountPunc (total count of the punctuations), SentimentScore[2] (reflects the tone, mood of the speaker displayed in the text), CountWord (total count of the words).

### 3.5 Word Embeddings

To deal with text in an ML problem, one way is to follow the regular ML routine to ask the right questions and cleverly figure out the hidden features, which are the essential differentiators, and train the model on those features. This model is likely to be more efficient in terms of complexity but requires good experience and knowledge in the underlying problem statement domain. Another approach is to try including all the features and let it be on the machine to figure out the differentiating ones. Although a more relaxed approach, this approach involves higher complexity. In NLP, we have made progress to represent full text in the form of vectors through word embeddings. Two of the most common approaches are TFIDF and Doc2Vec. It is but obvious that while transformation, we lose some of the features in the text as we are moving from a higher dimension to a lower-dimensional space. Nevertheless, if we are still able to retain the differentiating features, we will be able to train an ML model with a good level of accuracy. We have used both TFIDF and Doc2Vec in our experiments.

### 3.6 The ML models

There are many models available in the classification domain of ML. We have used SVM, KNN, Logistic Regression, and Random Forest in our experiments.

### 3.7 The ANN

Artificial Neural Networks (ANNs) have been the pioneers of Deep Learning and are the way to solve complicated, ill-defined problems of the Modern Deep Learning. We have also used ANN in our experiments.

### 3.8 The Hybrid Ensemble

The underlying motivation is the same as in any Ensemble technique. We intent to use the goodness of the classical ML models with the capabilities of the modern Deep Learning pioneer – the ANN.

### 3.9 The Architecture

We trained each model (from SVM, KNN, Logistic Regression, and Random Forest) using different combinations of the additional features we introduced, and also by using

---

[1] https://pypi.org/project/textstat/
[2] https://pypi.org/project/vaderSentiment/

TFIDF and Doc2Vec individually. We also trained the ANN using all the features we introduced earlier; we did not try any combination of features as ANN will automatically adjust weights to best use the given features. Finally, coming to the Hybrid Ensembles, we have four versions of these. In all of the hybrid models, we use only 60% of training data to train the classical models; then, we get their predictions on 40% of the remaining training data. Finally, we use this 40% of the remaining training data to train our hybrid model, an ANN. We experiment with each of these scenarios to see which one yields the best possible model and whether we can achieve any success with such hybrid ensembles. In Hybrid V1, we provide prediction vectors of the classical models (SVM, KNN, Logistic Regression, Random Forest) built using all the linguistic features along with all the linguistic features to ANN. This provides ANN with the opinion of other four statistical models along with the features to form its own opinion. In Hybrid V2, we only provide prediction vectors of classical models built using all linguistic features to ANN. In Hybrid V3 and V4, we feed prediction vectors of classical models trained on TFIDF and Doc2Vec, respectively, to ANN. Fig. 1. below depicts the hybrid architecture used. It is to be noted that the rightmost connection is ON only for Hybrid model V1 and is OFF for the other three Hybrid models (V2, V3 and V4).

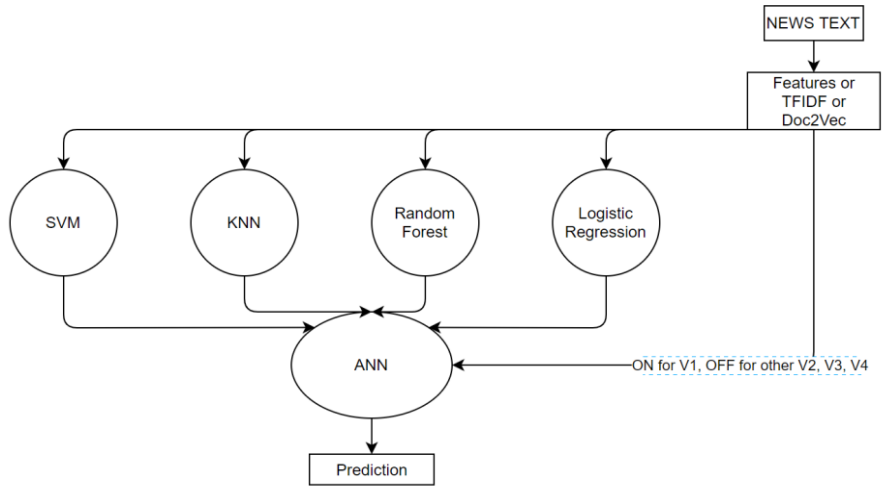

**Fig. 1.** The Hybrid Architecture

## 4 Experiments

We conducted experiments using each of the features, including the complete set for each classical model. We also used TFIDF and Doc2Vec individually on each classical model. The results have been captured in Tables 1 to 4. Finally, we experimented with ANN. The results of the experiment have been summarized in Table 5.

**Table 1.** SVM

| Features | Test | Validation |
|---|---|---|
| Readability | 56.35% | 52.02% |
| CountPunct | 56.27% | 51.48% |
| SentimentScore | 56.35% | 52.02% |
| CountWord | 56.35% | 53.89% |
| All Features | 56.43% | 53.19% |
| TFIDF | 61.72% | 61.29% |
| Doc2Vec | 56.35% | 52.02% |

**Table 2.** KNN

| Features | Test | Validation |
|---|---|---|
| Readability | 51.78% | 50.39% |
| CountPunct | 44.59% | 47.51% |
| SentimentScore | 53.20% | 52.02% |
| CountWord | 49.88% | 52.88% |
| All Features | 57.06% | 52.10% |
| TFIDF | 57.06% | 55.61% |
| Doc2Vec | 54.14% | 49.14% |

**Table 3.** Logistic Regression

| Features | Test | Validation |
|---|---|---|
| Readability | 56.35% | 52.02% |
| CountPunct | 56.35% | 52.02% |
| SentimentScore | 56.35% | 52.02% |
| CountWord | 56.35% | 52.18% |
| All Features | 56.35% | 53.27% |
| TFIDF | 61.64% | 59.50% |
| Doc2Vec | 56.35% | 52.02% |

**Table 4.** Random Forest

| Features | Test | Validation |
|---|---|---|
| Readability | 55.41% | 51.40% |
| CountPunct | 56.20% | 51.01% |
| SentimentScore | 55.85% | 50.62% |
| CountWord | 56.35% | 53.97% |
| All Features | 53.04% | 53.82% |
| TFIDF | 61.56% | 61.29% |
| Doc2Vec | 53.67% | 52.02% |

Table 5. ANN

| Features | Test | Validation |
|---|---|---|
| All Features | 55.85% | 54.05% |
| TFIDF | 57.77% | 57.17% |
| Doc2Vec | 56.35% | 52.02% |
| Hybrid V1 | 56.35% | 53.74% |
| Hybrid V2 | 56.59% | 53.89% |
| Hybrid V3 | 61.64% | 60.51% |
| Hybrid V4 | 56.35% | 52.02% |

## 5  Discussion

We can see that the accuracy was best in TFIDF with the model as SVM. TFIDF performed comparatively better in all the models. Even in the Hybrid models, the best performing model is Hybrid Ensemble V3, which uses the prediction vectors of classification models trained using TFIDF. One possible explanation for this is the higher dimension of data in TFIDF. We always lose some information in the data whenever we perform some processing to convert it form natural language to a machine understandable form, which is mathematical. It appears that TFIDF is able to retain some essential information which helps in separating fake news from the true news. This higher space, higher complexity, higher accuracy trend has been promising and is likely to work even in the future. If we come up with more elementary representations of textual data close to the natural language and train on a humongous corpus of data, we are likely to surpass the accuracy offered by TFIDF. All other variations in the feature vectors and models showed an ambivalent trend and did not promise much hope. Doc2Vec did not perform well since the average number of words in a statement was less, and the total number of statements also could not compensate for yielding essential relations between the words during the three layered shallow neural net training and failed to yield a meaningful Doc2Vec. Coming to the Hybrid Ensemble models, these results depict that we are unlikely to achieve success with such architecture models owing to several reasons. First, the Hybrid models were based on ANNs, but we did not have a considerable amount of quality data to train ANN well. Second, it may be that the Hybrid models are bound by the best performing model underneath and cannot perform better than it. This is just a hypothesis. Today, we have ensemble models that perform better than the underneath building models. Third, we experimented with a limited number of linguistic features, we may explore more features that may help us in improving the models, but it is going to be hard to determine which features are essential and why they are so. We could always use other NLP techniques such as N - grams and experiment with different Deep Learning models such as CNN and RNN variants like LSTM and get different results. Forth, even if we succeed in determining the differentiating features, our models will be successful only in a particular type of textual forms of fake news that play with those features. Whenever our model encounters something different, it will be tricked, and all predictions might be totally wrong. Fifth, even if we progress significantly in the linguistic feature department, we will still be left to deal with the expert knowledge department. We will have fake news which will be entirely equivalent to the true version in a grammatical sense but will be flawed due to the message being conveyed. Such news demands expert domain knowledge and cannot be dealt with by relying on pure linguistic features. To build a truly capable fake news detection model, we will have to incorporate domain knowledge in some form or the other. All code and data used in this paper is available at GitHub[3] for further experimentation.

## 6 Conclusion

Fake News Detection continues to be an interesting and complex domain of research. We discussed it is difficult to develop a truly robust machine learning model due to various factors. Although, there can be numerous ways to approach this problem such as linguistic approach, complex social networks (pattern of news generation, news flow pattern, source nodes, etc.), vector spaces (Word2Vec, Doc2Vec, TFIDF, etc.), etc. It appears as if the expert domain knowledge still is likely not replaceable by such static machine learning models. Dynamic algorithms which in effect use dynamic data as their ground truth could be a sound way forward in this area.

---

[3] https://github.com/singh-l/hybrrid_FN_dat_